\documentclass[a4paper, 10 pt, conference]{ieeeconf}
\IEEEoverridecommandlockouts
\usepackage{cite}
\usepackage{amsmath,amssymb,amsfonts}
\usepackage{algorithmic}
\usepackage{graphicx}
\usepackage{subfigure}
\usepackage{textcomp}
\usepackage{xcolor}
\usepackage{algorithm}
\usepackage{algorithmic}

\def\BibTeX{{\rm B\kern-.05em{\sc i\kern-.025em b}\kern-.08em
    T\kern-.1667em\lower.7ex\hbox{E}\kern-.125emX}}

\title{\LARGE \bf
Safe Reinforcement Learning for Autonomous Vehicles through Parallel Constrained Policy Optimization*
}

\author{Lu Wen$^{1,2}$, Jingliang Duan$^{1}$, Shengbo Eben Li$^{1*}$, Shaobing Xu$^{2}$, and Huei Peng$^{2}$ 
\thanks{*This study is supported by International Sci\&Tech Cooperation Program of China under 2019YFE0100200, and Beijing NSF with JQ18010. Special thanks should be given to TOYOTA for funding this study. L. Wen and J. Duan have equally contributed to this study. All correspondences should be sent to S. Li with email lisb04@gmail.com. }
\thanks{$^{1}$L. Wen, J. Duan and S. Li are with School of Vehicle and Mobility, Tsinghua University,
        Beijing, 100084, China. 
        {\tt\small Email: lisb04@gmail.com}}%
\thanks{$^{2}$L. Wen, S. Xu, H. Peng are with the Department of Mechanical Engineering,
        University of Michigan, Ann Arbor, 
        MI 48105, USA.   
        {\tt\small Email: lulwen@umich.edu}}%
}

\begin{document}

\maketitle
\thispagestyle{empty}
\pagestyle{empty}

\begin{abstract}
Reinforcement learning (RL) is attracting increasing interests in autonomous driving due to its potential to solve complex classification and control problems. However, existing RL algorithms are rarely applied to real vehicles for two predominant problems: behaviors are unexplainable, and they cannot guarantee safety under new scenarios. This paper presents a safe RL algorithm, called Parallel Constrained Policy Optimization (PCPO), for two autonomous driving tasks. PCPO extends today's common actor-critic architecture to a three-component learning framework, in which three neural networks are used to approximate the policy function, value function and a newly added risk function, respectively. Meanwhile, a trust region constraint is added to allow large update steps without breaking the monotonic improvement condition. To ensure the feasibility of safety constrained problems, synchronized parallel learners are employed to explore different state spaces, which accelerates learning and policy-update. The simulations of two scenarios for autonomous vehicles confirm we can ensure safety while achieving fast learning.
\end{abstract}

\section{Introduction}
Autonomous driving has the potential to improve safety and accessibility of ground vehicles. The approaches to design the control policy for autonomous vehicle generally fall into two categories: (1) rule-based methods and (2) learning-based methods. Because real-world driving can a lot of complexities, rule-based methods (e.g., the finite state mechanism) may not be able to handle all scenarios, and meanwhile pose significant burden on engineers to cover all possibilities \cite{duan2017driver}. The learning-based methods can imitate and learn from drivers' manipulation implicitly. In this study, we develop a learning-based method while making an important improvement.

Reinforcement learning (RL) has been actively studied for autonomous driving in recent years.  The goal of RL is to find policies to maximize the accumulated reward without reliance on labeled human driving data\cite{li2018RL}. Karavolos \textsl{et al}. (2013) first applied the vanilla Q-learning algorithm to efficiently train a driver in a racing game on the simulator TORCS\cite{karavolos2013q}. Silver \textsl{et al}. (2015) proposed a deep deterministic policy gradient (DDPG) algorithm by introducing deep learning into DPG algorithm\cite{silver2014deterministic}, which effectively solves problems in a continuous action space\cite{lillicrap2015continuous}. Wang \textsl{\textsl{et al}} (2018) successfully applied DDPG to an end-to-end policy learning for autonomous driving on TORCS\cite{wang2018deep}. Pan \textsl{et al}. (2017) built a new framework on the basis of A3C to train a self-driving vehicle by interacting with a synthesized real environment\cite{pan2017virtual}. Recently, Zhang \textsl{et al}. (2019) employed RL with model-based exploration for autonomous lane change decision-making on highways \cite{zhang2019discretionary}. Duan \textsl{et al}. (2020) employed RL with a hierarchical architecture for autonomous decision-making on highways \cite{duan2020hierarchical}\cite{duan2020distributional}.

To date, most RL methods have been developed on simulation platforms, with little work on real vehicles. A main reason is that the policy cannot be guaranteed to be safe, and the back-propagation-driven  process may lead to unforeseen accidents. Safety is the most basic requirement for autonomous vehicles, so a training process only look at reward, and not potential risk, is not acceptable. The notion of safe RL is defined in \cite{garcia2015comprehensive} as the process of learning policies that maximizes the expectation of accumulated return, while respecting security constraints in the learning and deployment process. More specifically, safe RL could be divided into being strictly safe and approximately safe. The algorithm developed in this paper is an approximately safe one. Approaches to solve this problem are categorized into two methods: (1) modifying the optimization criterion and (2) modifying the exploration process.

The method to modify the optimization criterion is to incorporate risk into the optimization objective, while the risk-neutral control neglects the variance in the probability distribution of rewards. 
We categorize these optimization criteria into four groups: maximin, risk-sensitive, constrained, and others. The maximin criterion considers a policy to be optimal if it has the maximum worst-case return. Gaskett (2003) considered the inherent uncertainty related to stochastic nature of the system, by proposing a new extension $\beta$-pessimistic term to Q-learning. Nilim and Ghaoui (2013) considered the uncertainty related to some of the parameters of the Markov decision process\cite{nilim2005robust}. Risk-sensitive criterion includes the notion of risk and return variance in the long term reward maximization objective. Geibel and Wysotzki (2005) transformed the optimization criterion into the probability of entering an error state\cite{geibel2005risk}. The constrained criterion ensures that the expectation of return is subject to one or more constraints. Castro \textsl{et al.} (2012) used a constrained criterion in which the variance of the return must not exceed a given threshold. More optimization criteria were explored to enforce safety as well\cite{tamar2012policy}. Mohammed \textsl{et al}. (2018) proposed a preemptive-shielding system, acting each time the learning agent is about to make a decision and providing a list of safe actions\cite{alshiekh2018safe}. But these methods have common drawbacks of turning overly pessimistic or computational intractability.

Modification of the exploration process can be categorized into two approaches: (1) incorporating external knowledge and (2) risk directed exploration. Incorporating external knowledge can provide initial knowledge to the agent, but it is not sufficient to prevent dangerous situations in later exploration. Siebel and Sommer (2007) used external knowledge as a form of population seeding in neuroevolution approaches\cite{siebel2007evolutionary}. Mohammed \textsl{et al}. (2018) introduced a new system named post-posed shielding. The shield monitors the agent's action and corrects them if the chosen action causes a violation. Risk directed exploration encourages the agent to explore controllable regions of environment by introducing risk metric as an exploration bonus. Garcia \textsl{et al}. (2012) successfully applied this approach to the helicopter hovering control in the RL Competition\cite{garcia2012safe}. Gehring and Precup (2013) defined a risk metric based on controllability\cite{gehring2013smart}. These methods have limited performance and are not always reliable because of their inability to detect risky situations for both early steps and long term.

A main contribution of this paper is to propose a safe RL algorithm, called Parallel Constrained Policy Optimization (PCPO), for autonomous driving policy. PCPO can ensure the policy is safe in the learning process and improve the convergence speed. PCPO considers a risk function and bounds the expected risk within predefined hard constraints. Meanwhile, a trust region constraint is added to allow large update step without breaking the monotonic improvement condition. The policy, value function and newly defined risk function are all approximated by neural networks. Secondly, synchronized parallel learners are employed to explore different state sub-space of system to reduce the correlation of sample sets, which increase the possibility of finding feasible states and accelerating the convergence speed. 

The rest of this paper is organized as follows: Section II introduces the PCPO algorithm, including the safety constraints, the trust region and the parallel learning framework. Section III presents two simulation studies for autonomous driving tasks: lane-keeping, and intersection crossing. Finally, we provide concluding remarks in section IV.
\section{Methodology}
\subsection{Preliminaries}
In this work, we formalize the RL problems into a Markov Decision Process (MDP) \cite{sutton2018reinforcement}. An infinite-horizon discounted MDP is defined by the tuple $\left(S,A,r,P,\rho_0,\gamma \right)$, where $S$ is the finite set of states, $A$ is the finite set of actions, $r:S\xrightarrow{}\mathbb{R}$ is the reward function, $P:S\times A\times S\xrightarrow{}\mathbb{R}$ is the transition probability distribution, $\rho_0:S\xrightarrow{}\mathbb{R}$ is the distribution of the initial state $s_0$, and $\gamma\in(0,1]$ is the discount factor.

For each state $s_t$ at time $t$, the expected accumulated return is defined as $R_t=\sum_{k=0}^{\infty}\gamma^k r_{t+k}$. The action is chosen according to a stochastic policy $\pi: S\xrightarrow{}P(A)$, with $\pi(a|s)$ denoting the probability of choosing action $a$ in state $s$.  The value function is $V^{\pi}(s)=\mathbb{E}_{\pi}\left[R_t|s_t=s\right]$, where $a_t\sim\pi(a_t|s_t)$, $s_{t+1}\sim P(s_{t+1}|s_t,a_t)$ for $t\geq0$, and the state-action value function is denoted as $Q^{\pi}(s,a)=\mathbb{E}_{\pi}\left[R_t|s_t=s,a_t=a\right]$.

RL aims to get the policy $\pi$ which maximizes the accumulated return in infinite horizon. Let $\eta(\pi)$ denote the objective function of policy update, it follows that $\eta(\pi)=\mathbb{E}_{\tau,\pi}\left[ \sum_{t=0}^{\infty}\gamma^t r(s_t)\right]$, where $\tau$ is a sequence of action-state: $\{s_0, a_0, s_1,\dots\}$. Obviously we can express $\eta(\pi)$ with $V^{\pi}$ and $Q^{\pi}$. Firstly, let's define the advantage function as
\begin{align}
\nonumber
\begin{split}
      A^{\pi}(s,a)&=Q^\pi(s,a)-V^\pi(s)\\
    &=\mathbb{E}_{s'}\left[r(s)+\gamma V^{\pi}(s')-V^{\pi}(s) \right].
\end{split}
\end{align}
So the advantage function $A^{\pi}$ and the objective function $\eta(\pi)$ satisfy
\begin{equation}
\nonumber
\eta(\pi)=\eta(\pi_{\text{old}})+\mathbb{E}_{\tau,\pi}\left[ \sum_{t=0}^{\infty}\gamma^t A^{\pi_{\text{old}}}(s_t,a_t)\right],
\end{equation}
where $\pi_{\text{old}}$ represents the old policy. It's sensible to use the state distribution corresponding to the old policy in replacement of that corresponding to policy $\pi$. In a large continuous state space, we can generally construct an estimator of the surrogate objective using importance sampling:
\begin{align}
\nonumber
\begin{split}
&\eta(\pi)\\
&=\eta(\pi_{\text{old}})+\sum_{s}\rho_{\pi}(s)\sum_a\pi(a|s)A^{\pi_{\text{old}}}(s,a)\\
&\approx \eta(\pi_{\text{old}})+\sum_{s}\rho_{\pi_{\text{old}}}(s)\sum_a\pi(a|s)A^{\pi_{\text{old}}}(s,a)\\
&=\eta(\pi_{\text{old}})+\mathbb{E}_{s,a\sim\pi_{\text{old}}}\left[\frac{\pi(a|s)}{\pi_{\text{old}}(a|s)}(Q^{\pi_{\text{old}}}(s,a)-V^{\pi_{\text{old}}}(s)) \right]\\
&=\eta(\pi_{\text{old}})+\mathop{\mathbb{E}}_{s,a\sim\pi_{\text{old}}}
     \left[\frac{\pi(a|s)}{\pi_{\text{old}}(a|s)}Q^{\pi_{\text{old}}}(s,a)\right] -\mathop{\mathbb{E}}_{s\sim\pi_{\text{old}}}\left[V^{\pi_{\text{old}}}(s)\right].
\end{split}
\end{align}
The goal of RL is to find the optimal $\pi$ that maximizes $\eta(\pi)$. Since $\eta(\pi_{\text{old}})$ and  $\mathbb{E}_{s\sim\pi_{\text{old}}}\left[V^{\pi_{\text{old}}}(s)\right]$ are independent of $\pi$, the policy optimization process can then be formulated as:
\begin{equation}
\nonumber
\pi=\arg \max\limits_{\pi \in \Pi} \eta(\pi)=\arg \max\limits_{\pi \in \Pi} \mathbb{E}_{s,a\sim\pi_{\text{old}}}\left[\frac{\pi(a|s)}{\pi_{\text{old}}(a|s)}Q^{\pi}(s,a) \right].
\end{equation}
For conciseness, we define the following surrogate objective function: \begin{equation}
\nonumber
J(\pi)=\mathbb{E}_{s,a\sim\pi_{\text{old}}}\left[\frac{\pi(a|s)}{\pi_{\text{old}}(a|s)}Q^{\pi_{\text{old}}}(s,a) \right].    
\end{equation}

\subsection{Actor-Critic}
RL methods usually employ an actor-critic (AC) architecture to approximate both the policy  and  value  function by iteratively solving the Bellman optimality equation based on generalized policy iteration framework. The AC architecture consists of two structures: the policy network (the so-called actor) and value approximation (the so-called critic)\cite{konda2000actor}. In this study, both the actor and critic are approximated by neural networks (NN), which directly map state to the probability distribution of action and expected cumulative return respectively. In this study, we adopt a stochastic policy, the output of which is the mean and standard deviation of the Gaussian distribution. We represent the policy network $\pi^{\boldsymbol{\theta}}(s)$ with parameters $\boldsymbol{\theta}$, and the state-action value network $Q^{\boldsymbol{\omega}}(s,a)$ with parameters $\boldsymbol{\omega}$.

In previous AC methods, the parameters $\boldsymbol{\omega}$ of the value network $Q^{\boldsymbol{\omega}}(s,a)$ are tuned by iteratively minimizing the following loss function
\begin{equation}
\nonumber
L_{\text{critic}}=(R_t - Q^{\boldsymbol{\omega}}(s_t,a_t))^2/2,
\end{equation} 
where $R_t - Q^{\boldsymbol{\omega}}(s_t,a_t)$ is usually called the temporal-difference (TD) error. The expected accumulated reward $R_t$ is usually estimated in the forward view using the n-step return:
\begin{equation}
\label{eq.update_rule_V}
R_t=\sum_{k=0}^{n-1}\gamma^k r_{t+k} + Q^{\boldsymbol{\omega}}(s_{t+n},a_{t+n}).
\end{equation} 
Then the specific gradient update for the parameters $\boldsymbol{\omega}$ of the value network is
\begin{equation}
\nonumber
\text{d}\boldsymbol{\omega}=(R_t-Q^{\boldsymbol{\omega}}(s_t,a_t))\nabla_{\boldsymbol{\omega}}Q^{\boldsymbol{\omega}}(s_t,a_t).
\end{equation} 

The parameters $\theta$ of policy network $Q^{\boldsymbol{\omega}}(s,a)$ are updated to maximize the surrogate objective function $J(\pi^{\boldsymbol{\theta}})$. Therefore, the update gradient of policy parameters $\theta$ is

\begin{equation}
\nonumber
\text{d}\boldsymbol{\theta}=\nabla_{\boldsymbol{\theta}}J(\pi^{\boldsymbol{\theta}})=\nabla_{\boldsymbol{\theta}}\mathbb{E}_{s,a\sim\pi^{{\boldsymbol{\theta}}_{\text{old}}}}\left[\frac{{\pi}^{\boldsymbol{\theta}}(a|s)}{\pi^{{\boldsymbol{\theta}}_{\text{old}}}(a|s)}Q^{\pi^{{\boldsymbol{\theta}}_{\text{old}}}}(s,a) \right].   
\end{equation}

Any standard NN optimization methods can be used to update these two NNs, including stochastic gradient descent (SGD), RMSProp, Adam, etc. Taking the SGD method as an example, the updating rules of the value network and the policy network are:
\begin{equation}
\nonumber
\boldsymbol{\omega} \leftarrow \boldsymbol{\omega}_{\text{old}} +\alpha_a\text{d}\boldsymbol{\omega},\quad
\boldsymbol{\theta}\leftarrow \boldsymbol{\theta}_{\text{old}} +\alpha_p\text{d}\boldsymbol{\theta},
\end{equation}
where $\alpha_v$ and $\alpha_p$ denote the learning rate of the value and policy networks, respectively.

Noted that only when the policy learning rate $\alpha_p$ is small enough, can the objective function $J(\pi^{\boldsymbol{\theta}})$ be guaranteed to be monotonously improved throughout the learning process\cite{schulman2015trust}. However, small learning steps usually leads to slow convergence. Another disadvantage is that policy safety cannot be guaranteed during the learning process.

\subsection{Algorithm}
\subsubsection{Constrained Policy Optimization}\quad
\par Inspired by the study of Achiam \textsl{et al}. (2016), we introduce a policy security constraint based on the newly defined risk function to ensure agent security during the learning process. This method is called Constrained Policy Optimization (CPO). Besides the reward signal $r$, the vehicle will also observe a scalar risk signal $\Tilde{r}$ at each step. The risk signal $\Tilde{r}$ is usually designed by human experts, which is usually assigned a large value when the vehicle moves into an unsafe state. Similar to the definition of $R_t$ and $Q(s,a)$, the expected accumulated risk $\Tilde{R}_t$ and the risk function (also called the cost function) $\Tilde{Q}(s,a)$ are defined as $\Tilde{R}_t=\sum_{k=\infty}^{n-1}\gamma^k \Tilde{r}_{t+k}$ and $\Tilde{Q}^{\pi}(s,a)=\mathbb{E}_{\pi}\left[\Tilde{R}_t|s_t=s,a_t=a\right]$ respectively. 
Similar to $J(\pi)$, we define the objective of $\Tilde{Q}(s,a)$: 
\begin{equation}
\nonumber
\Tilde{J}(\pi)=\mathbb{E}_{s,a\sim\pi_{\text{old}}}\left[\frac{\pi(a|s)}{\pi_{\text{old}}(a|s)}{\Tilde{Q}}^{\pi_{\text{old}}}(s,a) \right].
\end{equation}
To ensure policy security, the risk function $\Tilde{J}(\pi)$ of policy $\pi$ should always be bounded above the safe bound $\delta$. So, the policy security constraint can be formulated as:
\begin{equation}
\nonumber
\Tilde{J}(\pi) \le \delta.
\end{equation}

\begin{figure}[!htbp]
\centerline{\includegraphics[width=0.48\textwidth]{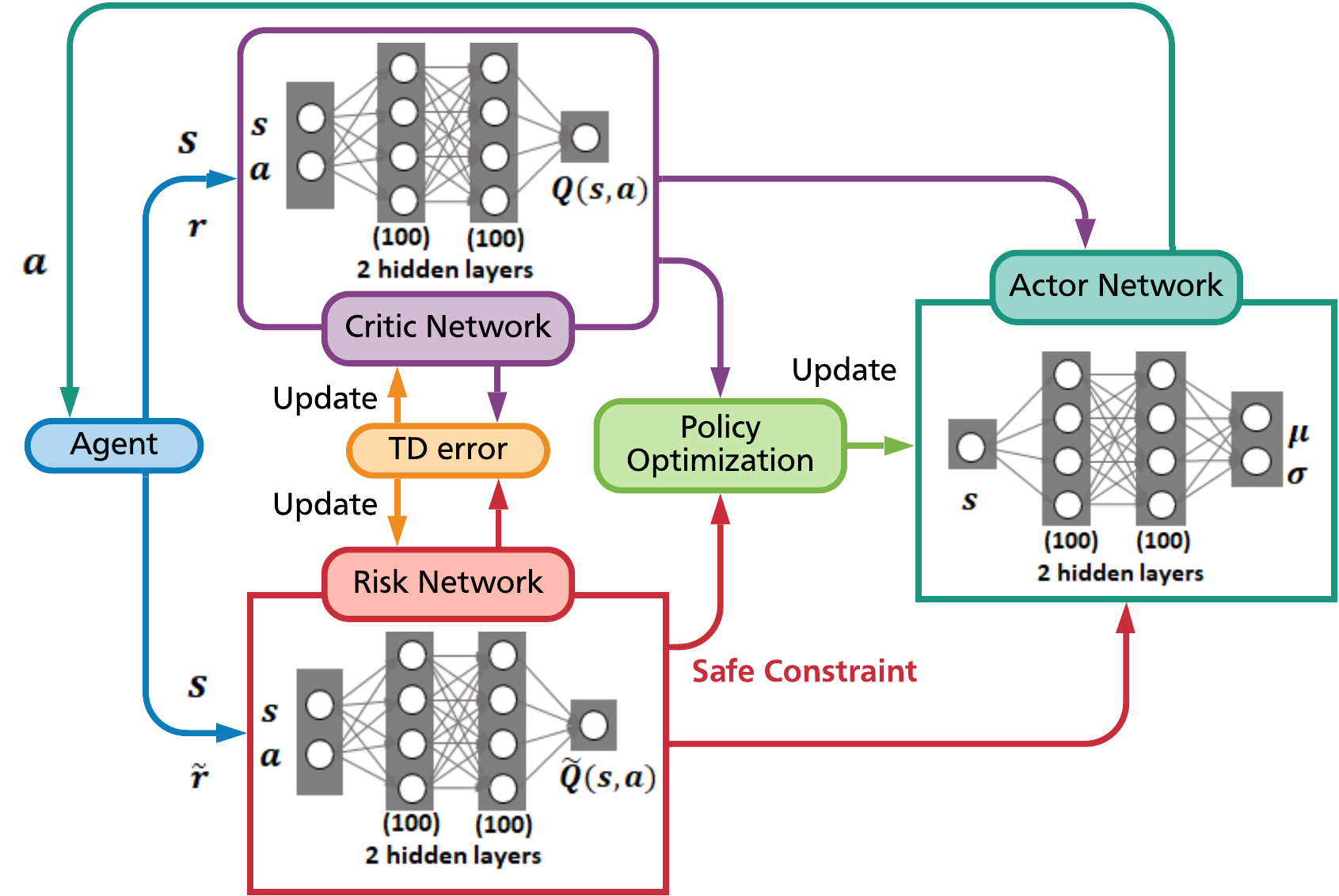}}
\caption{Actor-Critic-Risk architecture.}
\label{acc}
\end{figure}

In this study, the risk function  $\widetilde{Q}^{\boldsymbol{\phi}}(s,a)$ is also represented by a NN with parameters $\boldsymbol{\phi}$. The update method of the risk network $\widetilde{Q}^{\boldsymbol{\phi}}(s,a)$ is similar to the value network $Q^{\boldsymbol{\omega}}(s,a)$, which is omitted in this paper.
The policy, value and risk networks together constitute a new Actor-Critic-Risk (ACR) architecture (See Fig.\ref{acc}).

\subsubsection{Trust Region Constraint}\quad
\par
Since both $Q^{\boldsymbol{\omega}}(s,a)$ and $\widetilde{Q}^{\boldsymbol{\phi}}(s,a)$ are estimates, the monotonic improvement condition can be guaranteed only when the policy changes are not very large. Therefore, we add a constraint to avoid excessive policy update, so as to take relatively larger update steps without breaking the monotonic improvement guarantee inspired by \cite{schulman2015trust}. The policy constraint is described as: 

\begin{equation}
\nonumber
\mathbb{E}_{s\sim\pi^{{\boldsymbol{\theta}}_{\text{old}}}}\left[D_{\text{KL}}(\pi^{\boldsymbol{\theta}}(s),\pi^{{\boldsymbol{\theta}}_{\text{old}}}(s)) \right] \le \delta,
\end{equation}
where $\delta > 0$ is the corresponding step size bound and $D_{\text{KL}}$ is the Kullback-Leibler (KL) divergence, which is used to measure the difference between the new policy $\pi^{\boldsymbol{\theta}}(s)$ and old policy $\pi^{{\boldsymbol{\theta}}_{\text{old}}}(s)$. This constraint is also called the trust region constraint \cite{schulman2015trust}.  

Therefore, the policy optimization problem can be formulated as:
\begin{align}
\label{eq.opt}
\begin{split}
\boldsymbol{\theta}^{k+1}=&\arg \max\limits_{\boldsymbol{\theta}} J(\pi^{\boldsymbol{\theta}})\\
\mathbf{s.t.}\quad & \widetilde{J}(\pi^{\boldsymbol{\theta}})\leq d \\
&\mathbb{E}_{s\sim\pi^{{\boldsymbol{\theta}}_{\text{old}}}}\left[D_{\text{KL}}(\pi^{\boldsymbol{\theta}}(s),\pi^{{\boldsymbol{\theta}}_{\text{old}}}(s)) \right] \le \delta.
\end{split}
\end{align}
The optimization problem in \eqref{eq.opt} is equivalent to the following one, written in terms of expectations:
\begin{align}
\label{eq.opt2}
\begin{split}
\boldsymbol{\theta}^{k+1}=&\arg \max\limits_{\boldsymbol{\theta}} \mathbb{E}_{s,a\sim\pi^{{\boldsymbol{\theta}}_{\text{old}}}}\left[\frac{{\pi}^{\boldsymbol{\theta}}(a|s)}{\pi^{{\boldsymbol{\theta}}_{\text{old}}}(a|s)}Q^{\pi^{{\boldsymbol{\theta}}_{\text{old}}}}(s,a) \right]     \\
\mathbf{s.t.} \quad & \mathbb{E}_{s,a\sim\pi^{{\boldsymbol{\theta}}_{\text{old}}}}\left[\frac{{\pi}^{\boldsymbol{\theta}}(a|s)}{\pi^{{\boldsymbol{\theta}}_{\text{old}}}(a|s)}\widetilde{Q}^{\pi^{{\boldsymbol{\theta}}_{\text{old}}}}(s,a) \right]      \leq d\\
&\mathbb{E}_{s\sim\pi^{{\boldsymbol{\theta}}_{\text{old}}}}\left[D_{\text{KL}}(\pi^{\boldsymbol{\theta}}(s),\pi^{{\boldsymbol{\theta}}_{\text{old}}}(s)) \right] \le \delta.
\end{split}
\end{align}

The nonlinear constrained optimization problem formulated above is difficult to solve in practice due to the high-dimensional policy parameters $\boldsymbol{\theta}$. However, for small step sizes $\delta$, both the objective function $J(\pi^{\boldsymbol{\theta}})$ and risk function $\widetilde{J}(\pi^{\boldsymbol{\theta}})$ can be
approximated through linearizing around $\pi^{{\boldsymbol{\theta}}^k}$, and the
trust region constraint can also be well-approximated by the second order expansion at $\boldsymbol{\theta}=\boldsymbol{\theta}^k$. The local approximation to \eqref{eq.opt2} is:
\begin{align}
\label{eq.linear}
\begin{split}
\boldsymbol{\theta}^{k+1}=&\arg \max\limits_{\boldsymbol{\theta}} g^T(\boldsymbol{\theta}-\boldsymbol{\theta}^k)\\
\mathbf{s.t.}\quad & c+b^T(\boldsymbol{\theta}-\boldsymbol{\theta}^k)\leq 0 \\
&\quad   \frac{1}{2}(\boldsymbol{\theta}-\boldsymbol{\theta}^k)^TH(\boldsymbol{\theta}-\boldsymbol{\theta}^k)\leq \delta,   
\end{split}
\end{align}
where $g$ is the gradient of the objective $J(\pi^{\boldsymbol{\theta}})$, $b$ is the gradient of risk function $\widetilde{J}(\pi^{\boldsymbol{\theta}})$, $H$ stands for the Hessian of the KL-divergence $\mathbb{E}_{s\sim\pi^{{\boldsymbol{\theta}}_{\text{old}}}}\left[D_{\text{KL}}\right]$, and $c$ is defined as $c:= \widetilde{J}(\pi^k)-d$.

The optimization problem above is convex and can be solved efficiently using duality because $H$ is always positive semi-definite. We will assume it is positive-definite in the following. Denoting the Lagrange multipliers as $\lambda$ and $\nu$, the dual to the original problem can be expressed as:
\begin{equation}
\label{eq.dual_problem}
\max_{\begin{subarray}{c} \lambda>0\\ \nu\geq 0 \end{subarray}} \frac{-1}{2\lambda}(g^TH^{-1}g-2\nu^Tb^TH^{-1}g+\nu^Tb^TH^{-1}b\nu)+\nu^Tc-\lambda\delta.
\end{equation}
If \eqref{eq.linear} is feasible, and $\lambda^*,\nu^*$ are the optimal solution to \eqref{eq.dual_problem}, then the update rule for policy is:
\begin{equation}
\label{eq.update_rule}
\boldsymbol{\theta}^{k+1}=\boldsymbol{\theta}^k+\frac{1}{\lambda^*}H^{-1}(g-b\nu^*).
\end{equation}

\subsubsection{Parallel Constrained Policy Optimization}\quad\par

Sometimes we cannot find a feasible solution to \eqref{eq.linear}. The first reason is dangerous state, i.e., the risk function $\widetilde{J}(\pi^{\boldsymbol{\theta}})$ value is really high when the agent is in a dangerous state. Another reason is a dangerous action because CPO may take a bad update and produce an unsafe action due to approximation errors in \eqref{eq.linear}.  Achiam \textsl{et al}. handle this situation by proposing a recovery rule to decrease the constraint value \cite{achiam2017constrained}. The recovery rule is as follows:
\begin{equation}
\label{eq.recover}
\boldsymbol{\theta}^{k+1}=\boldsymbol{\theta}^k-\sqrt{\frac{2\delta}{b^TH^{-1}b}}H^{-1}b.
\end{equation}
After applying the recovery update, the constraint value is reduced so that the case turns feasible again. However, this recovery rule does not apply to dangerous state cases, because the policy $\pi^{\boldsymbol{\theta}^{k}}$ may act well in safe states. In this case, the adoption of this rule will result in slower convergence.

\begin{figure}[!htbp]
\centerline{\includegraphics[width=0.4\textwidth]{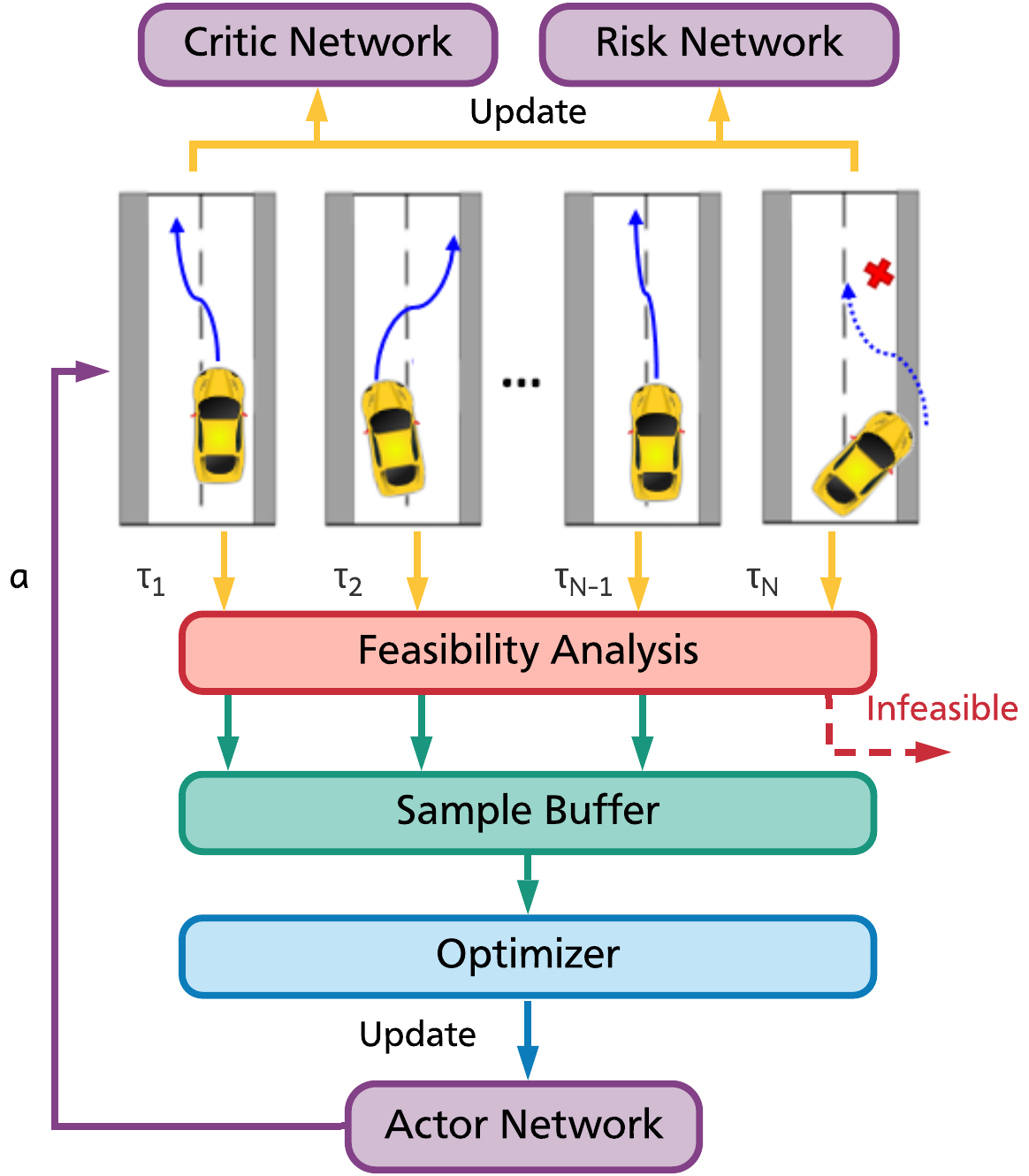}}
\caption{Parallel Constrained Policy Optimization structure.}
\label{parallel}
\end{figure}

To overcome this problem, we employ multiple agents to explore different state spaces in parallel. The general structure of the parallel algorithm is illustrated in Fig.\ref{parallel}. At each learning step, each agent synchronously generates samples based on the shared policy and uses its samples to solve \eqref{eq.linear}. We call samples $\tau_i$ collected by agent $i$ feasible if \eqref{eq.linear} is feasible. All samples from different agents will be used to update the value network and the risk network after each iteration. However, only feasible samples can be used to update the policy network. Whether a set of samples $\tau_i$ is feasible can be mathematically inferred with the following two indexes: $c_i$, and $e_i=\delta-\frac{{c_i}^2}{b^{T}H^{-1}b}$. It is easy to know by analysis that samples set $\tau_i$ is feasible only when $c_i>0$ and $e_i<0$.

If no feasible samples are collected in one learning step, \eqref{eq.recover} will be adopted to update the policy. An advantage of parallel agents learning is that it helps to reduce the correlation and increase the coverage of all collected samples, which leads to higher convergence speed and learning stability. The algorithm combining CPO and synchronous parallel agents learning is called Parallel CPO (PCPO) in this study. The pseudo code for PCPO algorithm is given below:

\begin{algorithm}
\caption{Parallel Constrained Policy Optimization}
\label{alg:A}
\begin{algorithmic}
{
\REQUIRE~~\\ Initial with arbitrary $\boldsymbol{\theta}$, $\boldsymbol{\omega}$ and $\boldsymbol{\phi}$ and state $s_0 \in S$
\ENSURE{}
\FOR{$k=1,2,\dots,n$}
\STATE Explore samples set $\tau=\{s\} \sim \pi(\boldsymbol{\theta}^k)$
\STATE Update the Value Network with $\text{d}\boldsymbol{\omega}$ in \eqref{eq.update_rule_V}
\STATE Update Risk Network with: \\
$\quad \quad \text{d}\boldsymbol{\phi}=(\widetilde{R}_t-\widetilde{Q}^{\boldsymbol{\phi}}(s_t,a_t))\nabla_{\boldsymbol{\phi}}{\widetilde{Q}}^{\boldsymbol{\phi}}(s_t,a_t)$
\STATE Estimate $g,b,H,c$ in \eqref{eq.linear} with $\mathcal{\tau}$ 
\STATE Store feasible $\mathcal{\tau}$ in buffer $D$
\ENDFOR
\IF{$D \not= \varnothing$}
\STATE Solve \eqref{eq.dual_problem} for $\lambda^*, \nu^*$ 
\STATE Update policy network using \eqref{eq.update_rule}
\ELSE
\STATE Recovery policy using \eqref{eq.recover}
\ENDIF
}
\end{algorithmic}
\end{algorithm}

\section{Experiments and Results}
We implement the PCPO algorithm to design autonomous driving functions. Two experiments were studied: the first one is a single-vehicle lane-keeping task, the second is a multi-vehicle interacting at an intersection.

\subsection{Lane keeping}
\subsubsection{Problem Description}
\begin{figure}[!htbp]
\centerline{\includegraphics[width=0.35\textwidth]{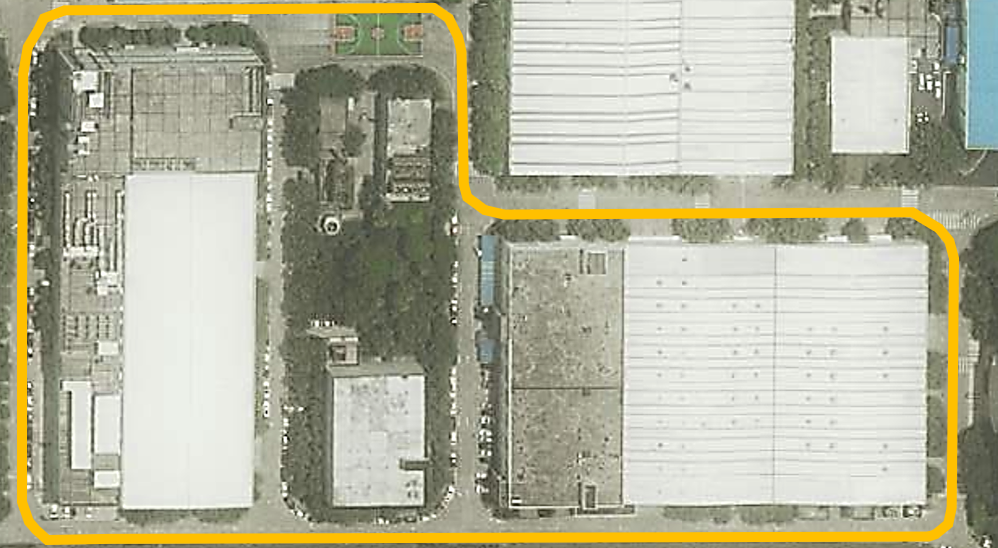}}
\caption{Test field map for the lane-keeping experiment.}
\label{map}
\end{figure}

The goal of the experiment is to keep the car as close to the center of the lane as possible while not deviating from the road throughout the learning process. The test road used in this experiment is a closed loop with a width of 3 m, which is shown in Fig. \ref{map}. The road position and direction information have been acquired by GPS every 0.015 meters. 

The state space of the lane-keeping task is represented by a tuple $S=\{d[\text{m}], \beta[\text{rad}]\}$, where $d$ denotes the relative lateral distance between the host vehicle and the lane center-line, and $\beta$ denotes the angle between the vehicle's heading angle and the tangent direction of current trajectory. Each parallel car-learner is initialized at a random position of this road. In this experiment, we only focus on the lateral control of the vehicle, and assume that the self-driving vehicle travels at a constant speed of 50 km/h. The action space is denoted by $A=\{\delta\text{[rad]}\}$, $\delta$ referring to the front wheel angle and $\delta\in [-\frac{\pi}{4},\frac{\pi}{4}]$. Given an action signal, the vehicle will move according to a two-degree bicycle dynamic model \cite{lee2018synthesis}.

The reward function is defined as follows:
$$r=-\frac{100}{9}d^2 - \beta^2.$$
Besides, the vehicle gets a risk of 100 if it leaves the lane boundary.

\begin{figure*}[!htbp]
\centering
\subfigure[Car-learner 1]{
\label{car1}
\includegraphics[width=0.22\textwidth]{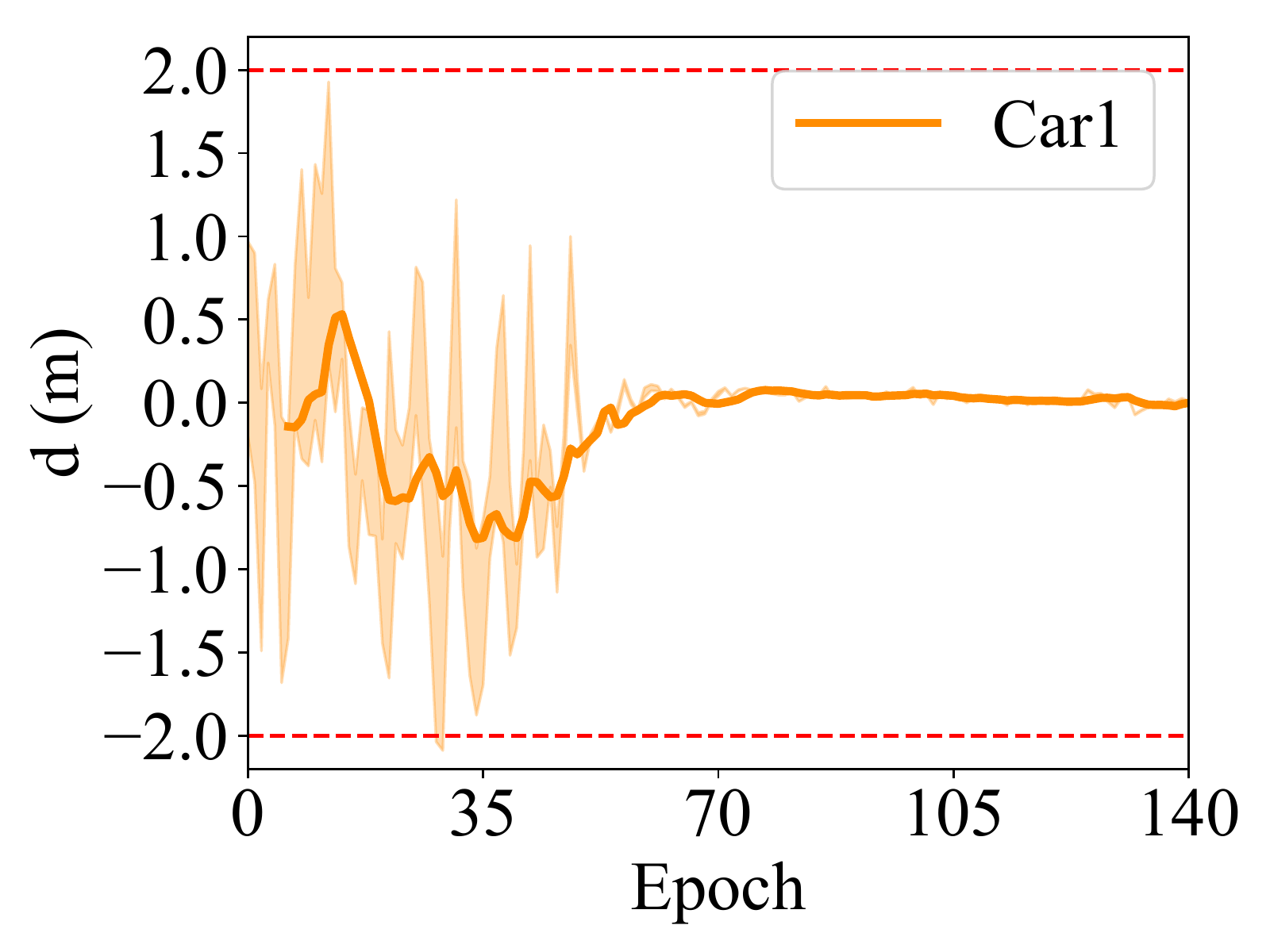}}
\subfigure[Car-learner 2]{
\label{car2}
\includegraphics[width=0.22\textwidth]{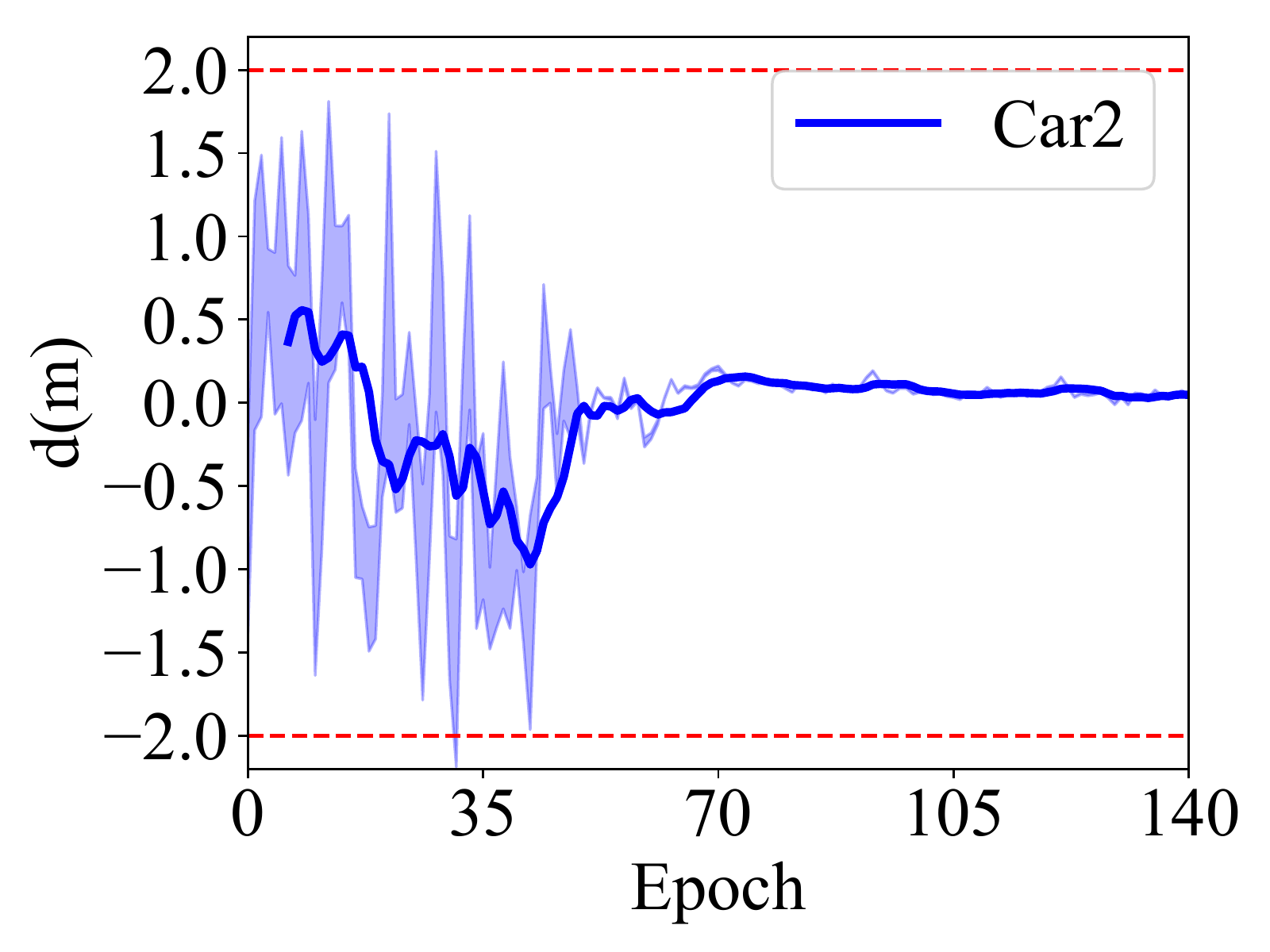}}
\subfigure[Car-learner 3]{
\label{car3}
\includegraphics[width=0.22\textwidth]{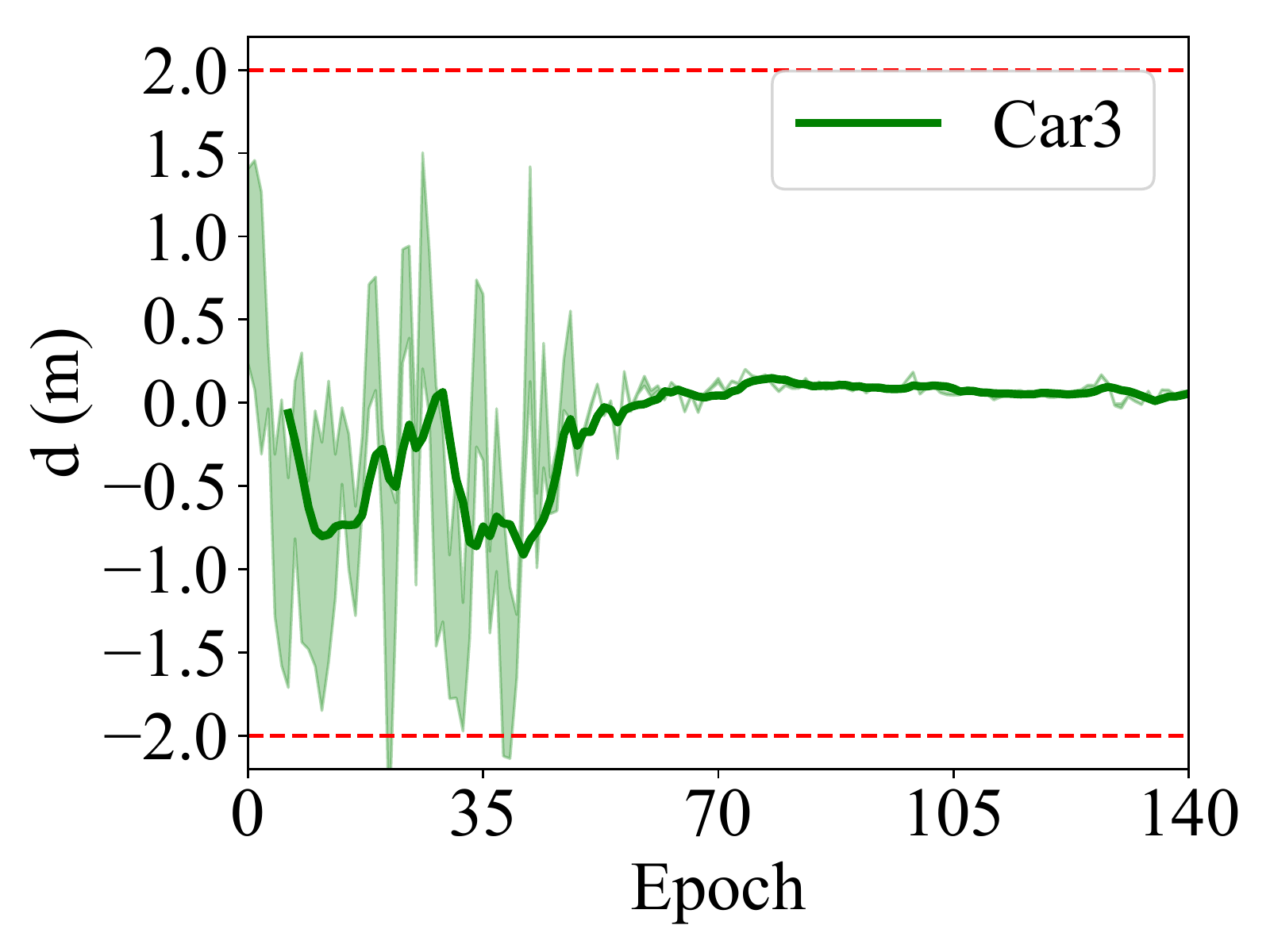}}
\subfigure[Car-learner 4]{
\label{car4}
\includegraphics[width=0.22\textwidth]{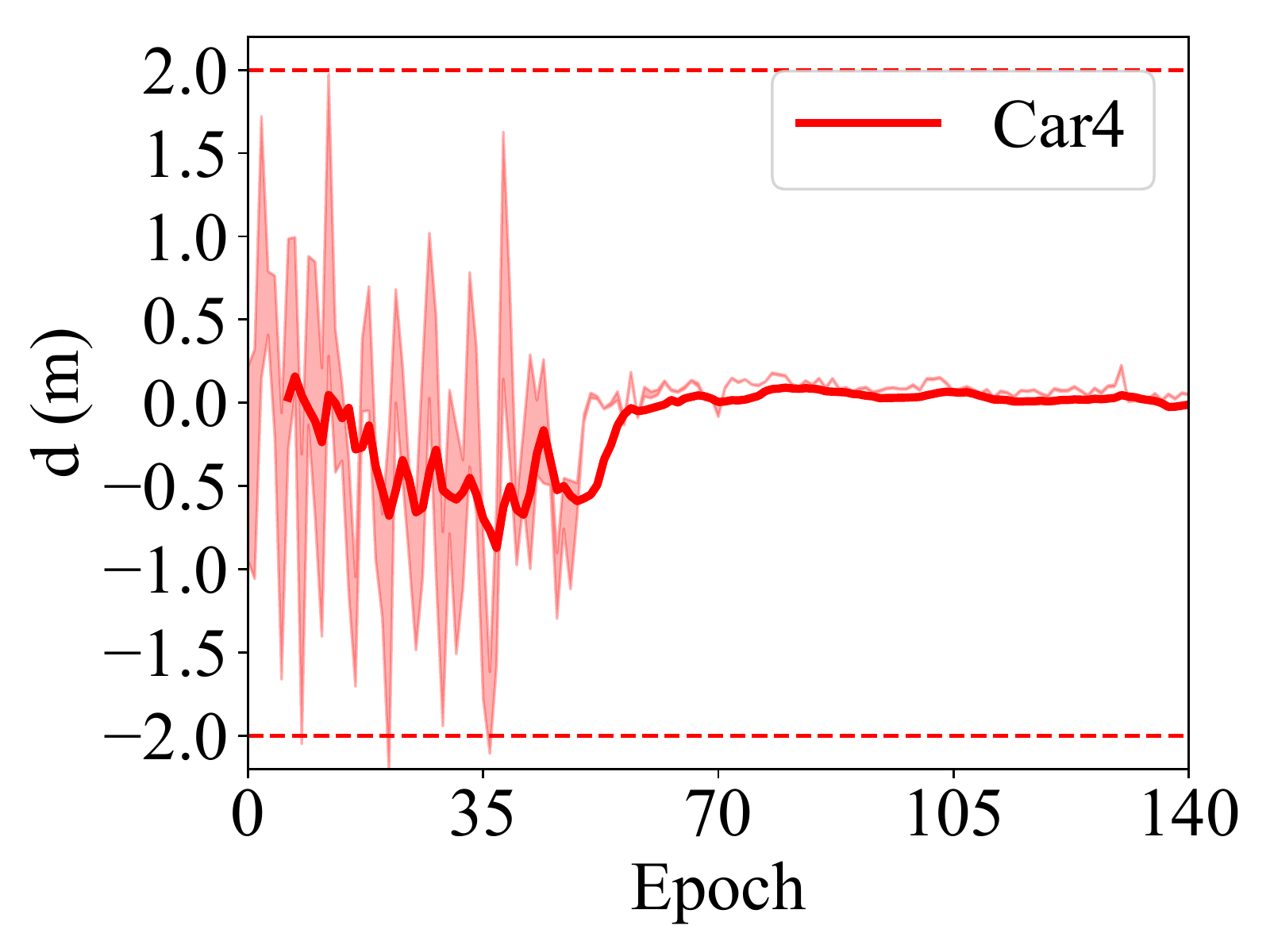}}
\caption{Lane-keeping experiment. Four parallel learning agents trained with the PCPO algorithm. The solid lines correspond to the mean and the shaded regions are from the maximum and minimum values of 5 runs. The red dash lines show the lane boundaries.}
\label{agents}
\end{figure*}

\begin{figure*}[!htbp]
\centering
\subfigure[Lateral Deviation Comparison]{
    \label{r1}
    \includegraphics[width=0.25\textwidth]{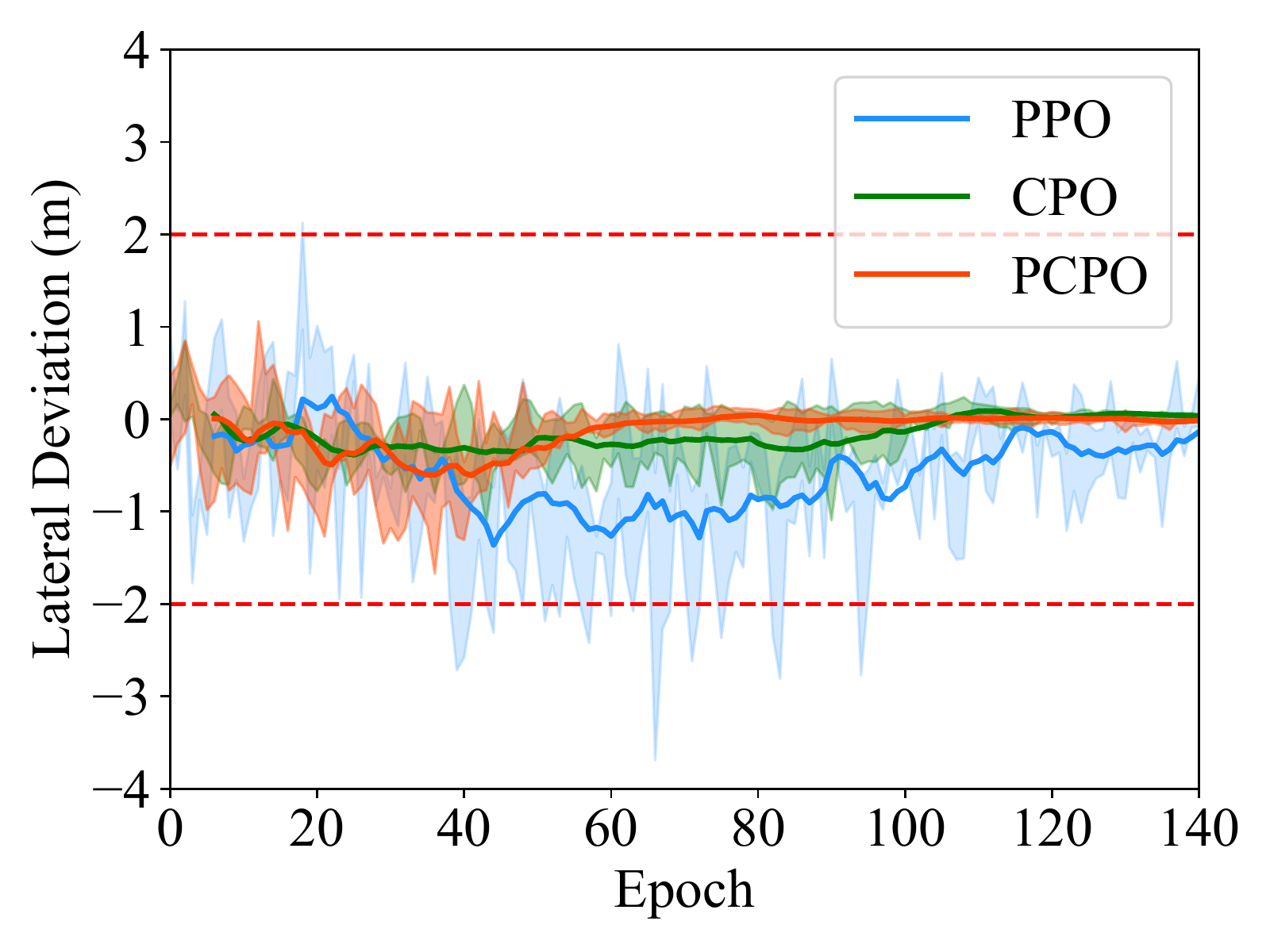}}
\subfigure[Risk Comparison]{
    \label{r2}
    \includegraphics[width=0.25\textwidth]{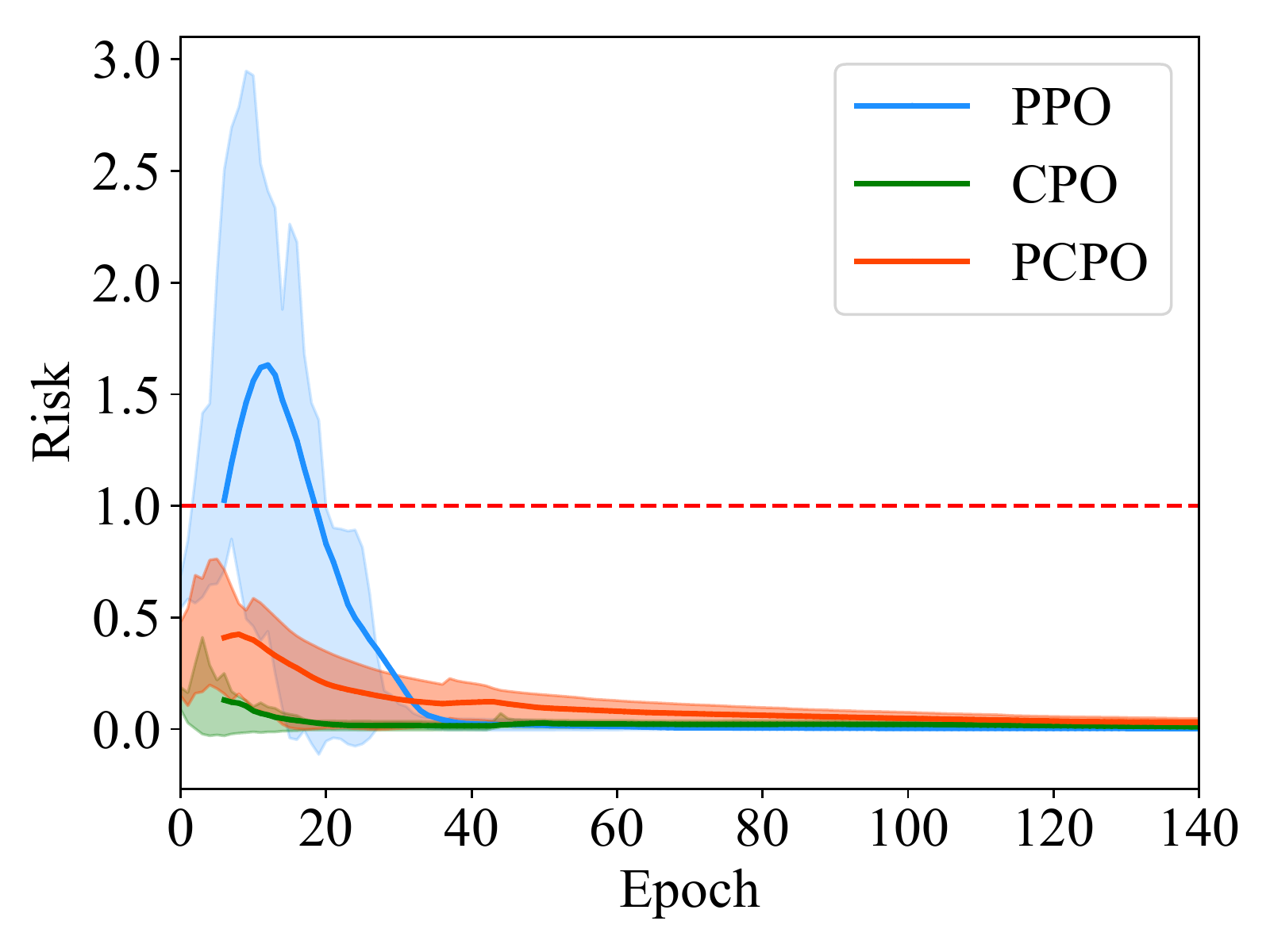}}
\subfigure[Return Comparison]{
    \label{r3}
    \includegraphics[width=0.25\textwidth]{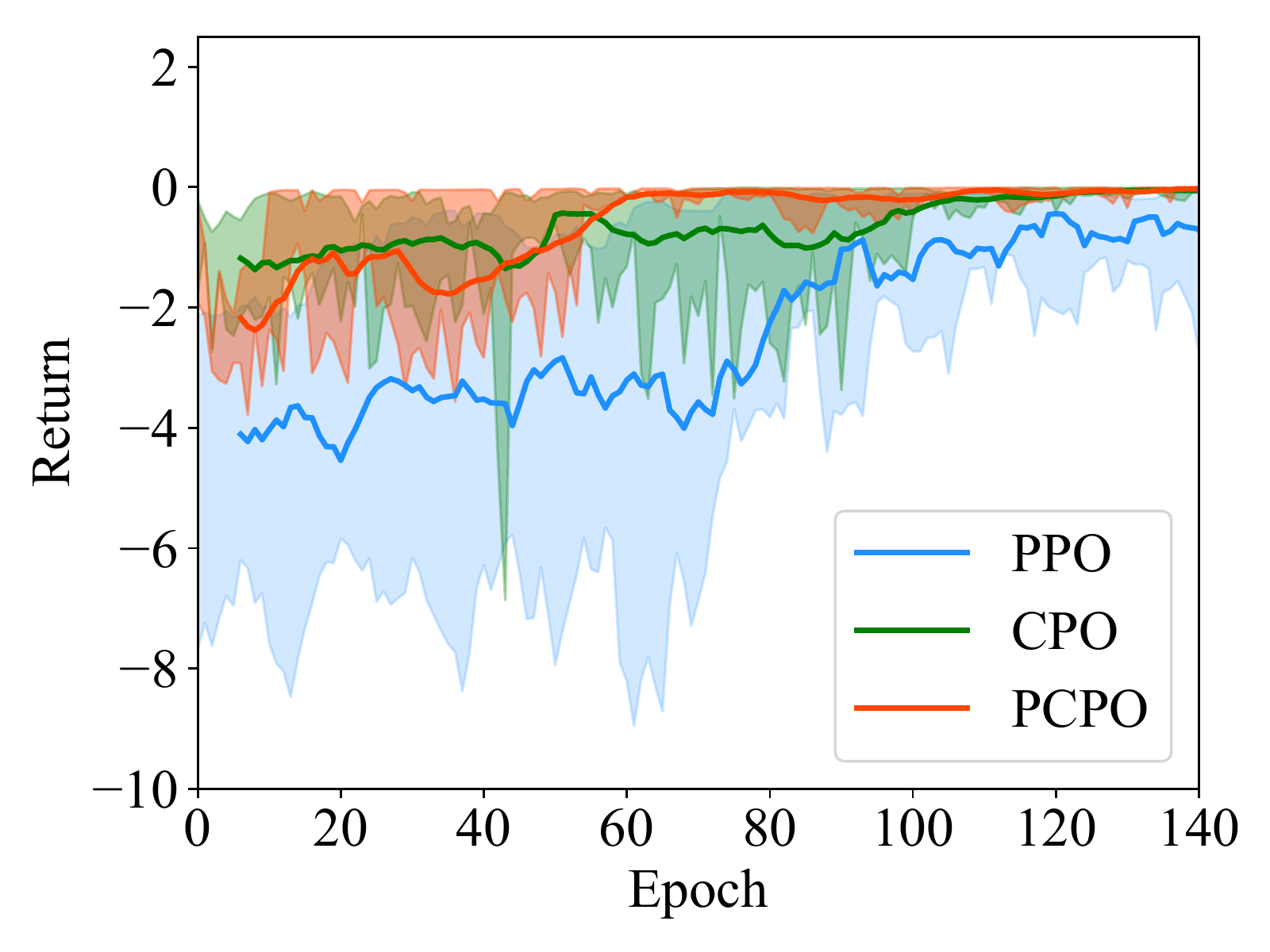}}
\caption{Learning curves comparison of the lane-keeping experiment. The safe limit ($d$) is set as 1, and the boundary delta ($\delta$) is set as $10^{-3}$. The red dash line stands for the safe limit. The solid lines correspond to the mean and the shaded regions correspond to standard deviation over 5 runs. This figure style is also applied in Fig.\ref{exp2-c}, Fig.\ref{exp2-r}.}
\label{results}
\end{figure*}

    \subsubsection{Algorithm Details}
    We employ four parallel cars to explore different state spaces and learn the sharing policy synchronously. Each car explores 16 steps at each iteration to form a sample set. Each epoch contains 25 episodes. The discount factor $\gamma = 0.95$. We learn the NN parameters with a learning rate of $10^{-3}$ for value and risk networks, while $10^{-4}$ for the policy network. For each NN, the input layer is composed of the states followed by 2 fully-connected layers with 100 hidden units for each layer. We use exponential linear units (ELUs) for hidden layers. Both the output layers of the value and risk networks are fully-connected linear layers with one scalar output. However, policy network has 2 outputs: 1) $\mu$ with activate function tanh, and 2) $\sigma$ with activate function softplus.
    \subsubsection{Results}
    Fig. \ref{agents} shows the evolution of the average lateral deviation of the four cars in 5 different runs during PCPO learning. Obviously, all parallel cars stay inside lanes throughout the learning process, while the deviation of each car quickly drops in about 70 epochs. This demonstrates PCPO's ability to ensure policy security during learning process while quickly converge to optimum.
    
    We compare the PCPO algorithm with two other RL algorithms, CPO and PPO. We used the same NN architecture and hyper-parameters for all three algorithms. Noted that, we use clipped surrogate objective function with $\epsilon = 0.2$ in the PPO experiment. Fig. \ref{r1} shows the training performance of all algorithms in 5 different runs. We can see that all three methods can eventually learn a safe lane-keeping policy, however, the vehicle deviates from the lane multiple times during the learning process of the PPO. Besides, in this task, PCPO improves learning speed by approximately 35\% compared to CPO, and by more than 70\% compared to PPO. Fig. \ref{r2} and \ref{r3} respectively plot the average risk value and return of all three algorithms. Fig. \ref{r2} indicates that only CPO and PCPO kept the expected risk value below the predefined risk limit. Fig. \ref{r3} indicates that PCPO can learn to obtain a relatively optimal policy while ensuring security and efficiency.

\subsection{Decision-making of multi-vehicles at an intersection}
    \subsubsection{Problem Description}
    In this experiment, an unsignalized intersection is chosen as the simulation scenario, where each direction is a bidirectional single carriageway. We consider three vehicles in the intersection, the trajectories of which are pre-assigned and fixed. We randomly initialize the velocity and position of each vehicle along its track, then implement algorithms to learn a centralized policy for the three vehicles to pass through the crossing as fast as possible and without collision. 
    \begin{figure}[!htbp]
        \setlength{\belowcaptionskip}{-2cm}
        \centerline{\includegraphics[width=0.3\textwidth]{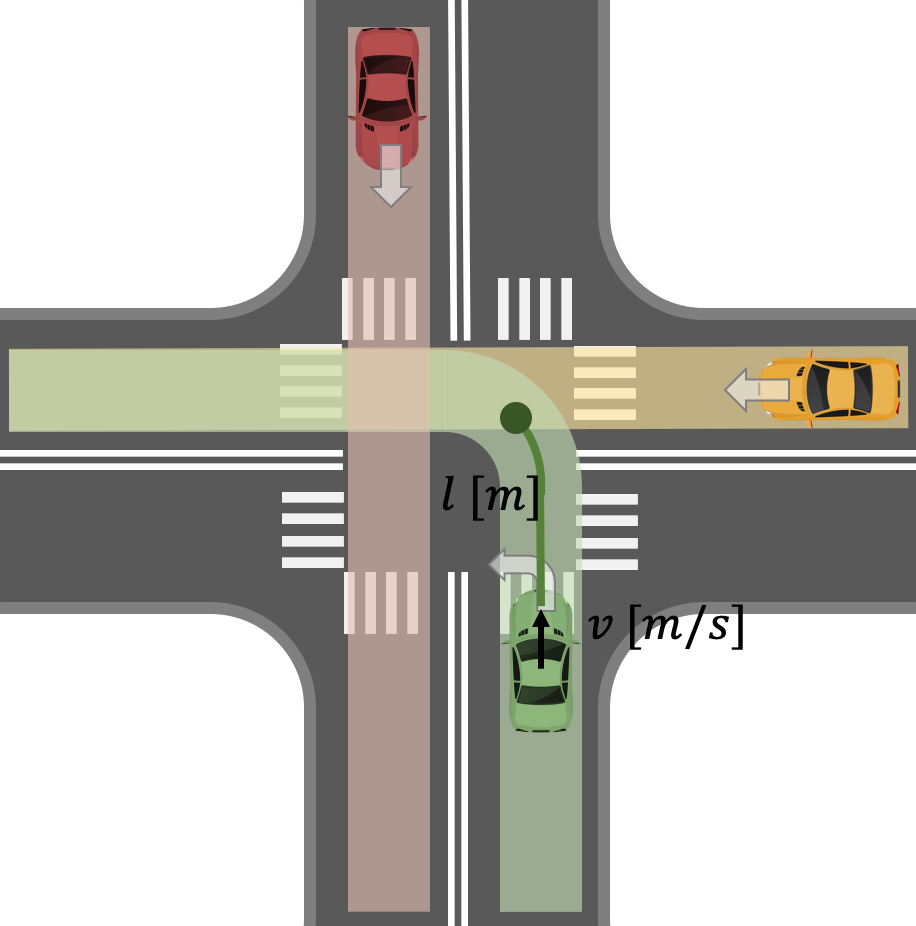}}
        \caption{The intersection crossing experiment. The colored bars show the pre-assigned trajectories. The green dot is an example to illustrate the middle point of the green vehicle's trajectory.}
        \label{exp2} 
    \end{figure}

    As Fig. \ref{exp2} shows, the state space is represented by a tuple $S=\{l_1,v_1,l_2,v_2,l_3,v_3\}$, where $l$ denotes the distance of the vehicle to the middle point of its track, and $v\in [6,14]$ (m/s) denotes the velocity. As the trajectory of each vehicle is fixed, the action space consists of the accelerations of the three cars $A=\{a_1,a_2,a_3\}$, where $a\in [-3,3]$ (m/$\text{s}^2$). The agents receive a reward of 10 for every passing vehicle, as well as a reward of -1 for every time step and an additional reward of 10 for terminal success. The agents are given a risk of 50 when a collision occurs.

    \subsubsection{Results}
    \begin{figure}[!htbp]
        \centerline{\includegraphics[width=0.4\textwidth]{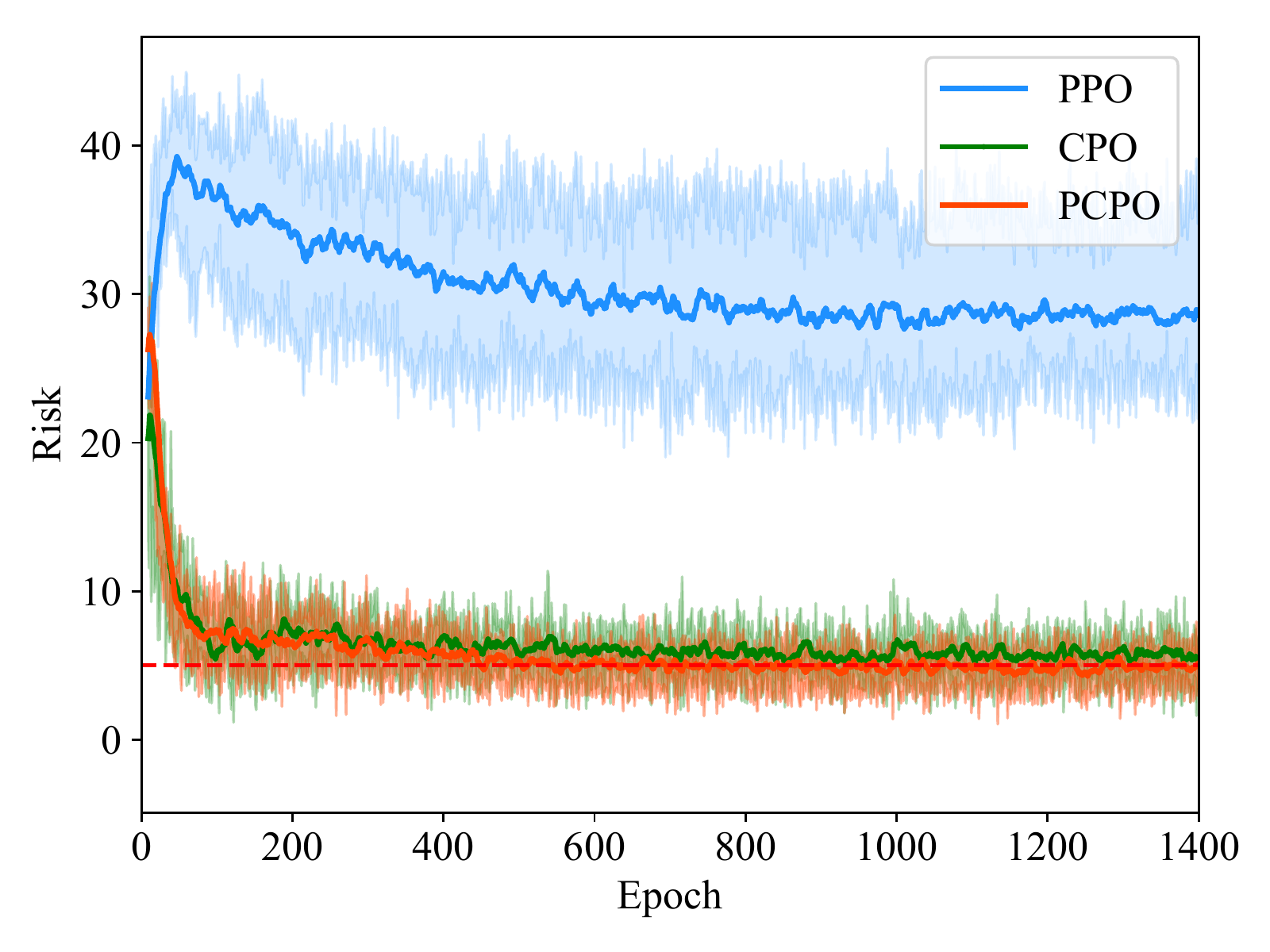}}
        \caption{Risk comparison of the crossing experiment. The safe limit ($d$) is set as 5, and the boundary delta ($\delta$) is set as $10^{-3}$. The closer to the limit is better.} 
        \label{exp2-c}  
    \end{figure}
    We share the same comparing algorithm types and settings, neural network structure, hype- parameters, etc. with the Lane Keeping experiment. From Fig. \ref{exp2-c} we can see that, risks of PCPO and CPO are monotonically reduced and kept around the safe bound throughout the process, which validates PCPO and CPO's better performance of guaranteeing safety during the learning process. Because observing safety constraints and getting high rewards are adversarial, specifically crossing and safety are conflicting in some way, thus the closer to the limit, the better.
    \begin{figure}[!htbp]
        \centerline{\includegraphics[width=0.4\textwidth]{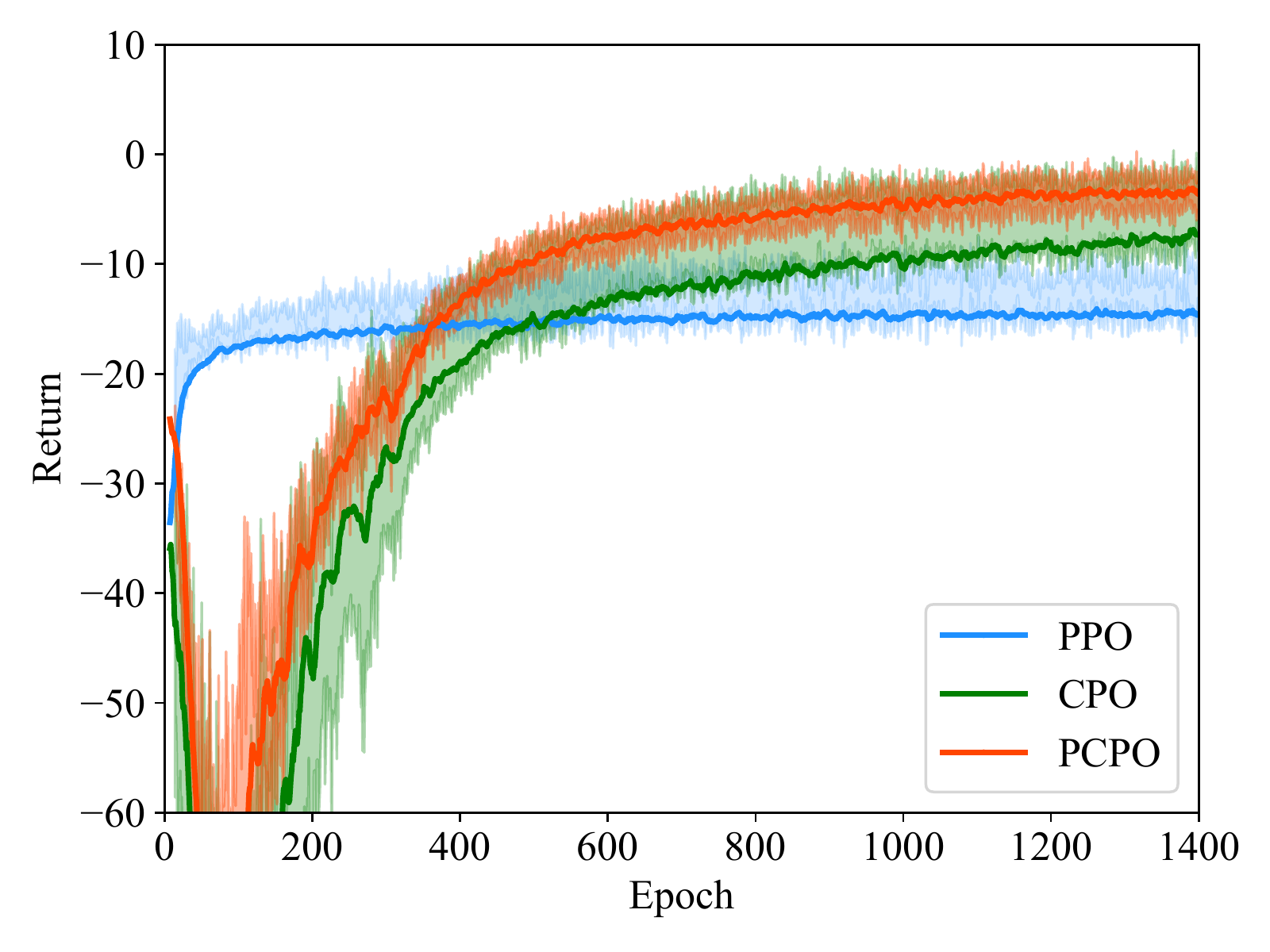}}
        \caption{Return comparison of the crossing experiment. The theoretical optimal return is 0.}
        \label{exp2-r}  
    \end{figure}
    In addition, from Fig. \ref{exp2-r} we can observe that PCPO has better learning performances in both learning speed and convergent optimality. The PCPO algorithm converges to 0 (theoretical highest return) after approximate 800 epochs' learning, in comparison the CPO algorithm reaches less optimal performance of -5 and learns slower, not converged until 1400 epochs (75\% slower than PCPO); the PPO algorithm seems to get the highest promotion in the initial epochs, however combining its risk of 30, much higher than the limit at 5, and its return learning curves, converging to the value of -15, we can infer that it converges to a sub-optimum, which is even an unsafe policy.
    
\section{Conclusion}

This paper presents a safe RL algorithm, called Parallel Constrained Policy Optimization (PCPO), for autonomous driving tasks.  PCPO is formulated to be a constrained optimization problem, in which an expected risk function bounded above by risk limit is introduced to guarantee policy safety. This algorithm extends today's actor-critic architecture to a three-component learning framework, in which three fully connected NNs are used to approximate policy, value function and newly defined risk function, respectively. Besides, a trust region constraint is added to allow large policy update without breaking the monotonic improvement condition. A synchronized parallel learning strategy is developed  to accelerate exploration and improve the possibility of achieving the optimal solution. We apply our algorithm to two tasks: one-vehicle lane-keeping and multi-vehicles at a crossing.
The experimental results show the contributions of the PCPO algorithms in solving autonomous driving problems from the following aspects:
\begin{itemize}
    \item It can guarantee safety constraints during the learning process for general autonomous driving tasks;
\item It has higher learning speed and higher data efficiency;
\item It also has more possibility to prevent learning agents from being stuck at sub-optima, or at least to a safe sub-optimal policy.
\end{itemize}

\bibliographystyle{ieeetr}
\bibliography{reference}
\end{document}